\pdfoutput=1

\documentclass[11pt]{article}

\usepackage[]{acl}

\usepackage[T1]{fontenc}

\usepackage[utf8]{inputenc}
\usepackage{rotating}

\usepackage{microtype}
\usepackage{times}
\usepackage{latexsym}
\usepackage{graphicx}
\usepackage{makecell}
\usepackage{tabularray}
\usepackage{multirow}
\usepackage{inconsolata}
\usepackage{enumitem}
\usepackage{booktabs}
\usepackage{float}
\usepackage{csquotes}
\usepackage{amssymb}  
\usepackage{changepage}

\usepackage{listings}
\usepackage{xcolor}

\lstdefinestyle{json}{
    basicstyle=\footnotesize\ttfamily,
    commentstyle=\color{gray},
    stringstyle=\color{blue},
    keywordstyle=\color{red},
    numbers=none,
    showstringspaces=false,
    breaklines=true,
    captionpos=b,
    tabsize=2,
    language=Python 
}
\lstdefinestyle{json_tab}{
    basicstyle=\scriptsize\ttfamily,
    commentstyle=\color{gray},
    stringstyle=\color{blue},
    keywordstyle=\color{red},
    numbers=none,
    showstringspaces=false,
    breaklines=true,
    breakindent=0pt,
    captionpos=b,
    tabsize=2,
    language=Python 
}
\lstdefinestyle{text_tt}{
    basicstyle=\ttfamily\scriptsize,
    commentstyle=\color{gray},
    stringstyle=\color{blue},
    keywordstyle=\color{red},
    numbers=none,
    breakindent=0pt,
    frame=none,
    breaklines=true,
    language=TeX,
    moredelim=**[is][\bfseries]{@}{@}
}
\lstdefinestyle{text_rm}{
    basicstyle=\ttfamily\small,
    commentstyle=\color{gray},
    stringstyle=\color{blue},
    keywordstyle=\color{red},
    numbers=none,
    breakindent=0pt,
    frame=none,
    breaklines=true,
    language=TeX,
    moredelim=**[is][\bfseries]{@}{@}
}

\makeatletter
\def\hlinewd#1{%
\noalign{\ifnum0=`}\fi\hrule \@height #1 %
\futurelet\reserved@a\@xhline}
\makeatother
\title{TOAD: Task-Oriented Automatic Dialogs with Diverse Response Styles}

\author{
\textbf{Yinhong Liu}$^{\spadesuit \heartsuit }$\thanks{\hspace{1mm} Work done during an internship at Apple.} \quad \quad
\textbf{Yimai Fang}$^\heartsuit$ \quad \quad
\textbf{David Vandyke}$^\heartsuit$ \quad \quad
\textbf{Nigel Collier}$^\spadesuit$ \quad \quad
\\
$^\spadesuit$Language Technology Lab, University of Cambridge \\
$^\heartsuit$Apple\\
{\tt\{yl535,nhc30\}@cam.ac.uk}\\
{\tt\{yimai\_fang,dvandyke\}@apple.com}
 }

\begin{document}
\maketitle
\begin{abstract}
In light of recent advances in large language models~(LLMs), the expectations for the next generation of virtual assistants include enhanced naturalness and adaptability across diverse usage scenarios.
However, the creation of high-quality annotated data for Task-Oriented Dialog~(TOD) is recognized to be slow and costly.
To address these challenges, we introduce Task-Oriented Automatic Dialogs~(TOAD), a novel and scalable TOD dataset along with its automatic generation pipeline.
The TOAD dataset simulates realistic app context interaction and provide a variety of system response style options.
Two aspects of system response styles are considered, verbosity level and users' expression mirroring.
We benchmark TOAD on two response generation tasks, and the results show that modeling
more verbose responses or responses without user expression mirroring is more challenging.\footnote{The data and code are available at \url{https://github.com/apple/ml-toad}.}

\end{abstract}

\section{Introduction}
\label{sec:introduction}

Task-Oriented Dialog (TOD) stands as a fundamental task of machine intelligence~\citep{5447049}, involving goal-driven conversations between a human and a system to achieve specific tasks.
It has been used in training virtual assistant systems for various real-world applications, including mobile phones~\citep{hoy2018alexa}, virtual reality devices~(VR)~\citep{kottur-etal-2021-simmc}, smart home devices~\citep{duong-etal-2019-adaptable}, online shopping assistants~\citep{yan2017building}, and trip booking helpers~\citep{el-asri-etal-2017-frames}.

Advancements in developing TOD systems face a challenge posed by conflicting requirements~\citep{hu2023multi3woz}: the demand for extensive datasets clashes with the substantial time and financial investments~(months if not years) required for data collection~\citep{larson2022survey}.
Looking ahead to the next generation of virtual assistants, an expectation arises for them to adapt their response styles to different usage scenarios, e.g. if equipped with a screen, ensuring a sustained level of naturalness and communication efficiency and ultimately enhancing the user experience.
However, existing datasets lack consideration for \textbf{adaptive response styles} and neglect to simulate \textbf{interactions with app contexts} like calendars or alarms.

\begin{figure}
    \centering
    \includegraphics[width=\linewidth]{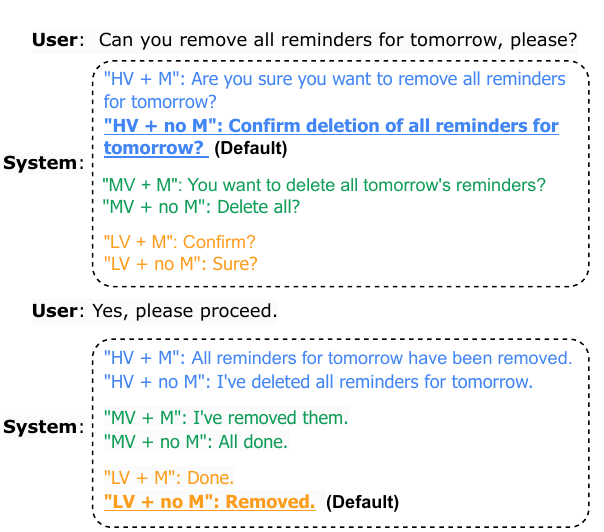}  
    \caption{A dialog example from TOAD, with all system response styles. LV, MV, HV and M stand for Low, Mid, High Verbosity and Mirroring. The underscored responses are selected as the default styles.}
    \label{fig:response style}
    \vspace{-5mm}
\end{figure}

\begin{table*}
\centering
\vspace{-2mm}
    \scalebox{0.95}{
        \begin{tabular}{lcccccc}
        \hline
                                        & MultiWoZ & PRSTO & ABCD & SGD     & STAR    & \textbf{TOAD} \\ \hline
        Number of dialogs               & 8,438    & 95,671& 8,034 & 16,142  & 5,820    & 8,087 \\
        Number of services       & 7        &34     & 30   & 16      & 13      &  11  \\
        Total number of turns           & 115,434  &-      &177,407& 329,964 & 127,833 & 37,678\\
        Average tokens / turn           & 13.1     &9.0    & 9.1  & 11.2    & 11.2    & 10.6 \\ \hline
        Context app interaction             &\textbf{\texttimes} & \checkmark& \textbf{\texttimes}& \textbf{\texttimes} & \textbf{\texttimes}&  \checkmark     \\
        Response style control          &\textbf{\texttimes} &\textbf{\texttimes} &\textbf{\texttimes} & \textbf{\texttimes} & \textbf{\texttimes} &  \checkmark     \\
        Highly automatic & \textbf{\texttimes} & \textbf{\texttimes} & \textbf{\texttimes} & \textbf{\texttimes} & \textbf{\texttimes} & \checkmark \\ \hline
        \end{tabular}
    }
\caption{Comparison of our TOAD dataset to other existing TOD datasets.}
\label{tab:comparison}
\vspace{-3mm}
\end{table*}

This paper presents Task-Oriented Automatic Dialog, TOAD, a highly automatic generated dataset, along with its generation pipeline, designed to address the challenges mentioned above.
In TOAD, we not only simulate realistic interactions with app context and diverse conversational phenomena, but also analyze two aspects of response style—\textbf{verbosity level} and \textbf{users' expressions mirroring}, aiming to enhance naturalness and adaptiveness for various usage scenarios.
As shown in the dialog example in Fig.~\ref{fig:response style}, for each turn of system utterance, we provide a spectrum of six responses style options.
The dataset is constructed using an automatic generation pipeline, leveraging the zero-shot generation capability of the latest Large Language Models~(LLMs), such as ChatGPT\footnote{https://openai.com/blog/chatgpt}.
Due to its automatic nature, the TOAD dataset is scalable in terms of data size and service coverage. 
To provide a benchmark for future studies, we establish reference scores for two TOD Natural Language Generation (NLG) tasks, evaluating a variety of baseline models. The results indicate that modeling more verbose or responses without user expression mirroring is more challenging.




The contributions of this work are three-fold:
1) A new TOD dataset with multiple response styles and realistic app context interaction, such as revising or deleting calendar events.
2) TOAD is the first that investigates the naturalness and adaptiveness of system response, providing insights into the dimensions of style that virtual assistants should consider.
3) An automatic TOD data generation pipeline for scalable, cost-effective expansion of data size and domains.




\section{Related Works}
\subsection{Task-Oriented Dialog Dataset}
To our best knowledge, TOAD stands out as the first highly automatically generated TOD dataset. A concurrent work, LUCID \citep{stacey2024lucid}, also adopts a similar LLM-driven approach for dialog generation.  All previous datasets relied on human annotations or paraphrasing, which makes the collection process costly and time-consuming.
Moreover, our dataset also admits all of the unique properties listed in Section~\ref{sec:introduction}.

Previous TOD datasets mainly fall into two categories based on how dialog utterances are collected: Machine-to-Machine (M2M) and Wizard-of-Oz (WOz).
In the M2M setup, proposed by \citet{shah2018building}, systems or schemas exhaustively simulate dialog with skeleton plots, represented in structured formats.
The dialog plots are realized into natural language by crowd workers.
A representative dataset is the SGD~\citep{rastogi2020towards}, which consists of dialogs across 16 domains.
They defined a list of meta information as schema, such as valid slots and supporting intents for each domain and utilize the slot-filling strategy to simulate the dialog plots. 
Another dataset, STAR~\citep{mosig2020star}, defined explicit ideal dialog flows for each domain and simulated realistic user behaviors such as small-talk interruptions.

Another line of TOD data collection setup, the WOz~\citep{kelley1984iterative}, employs crowd workers playing roles both of the user and the system, to directly produce utterances in a more improvised manner. 
The user is provided with an overall goal to achieve throughout the conversation, while the ``system'' needs to respond with access to a database based on the user's preference.
Such WOZ set-up has been successfully validated by WOz~\citep{wen2017network} and FRAMES~\citep{el2017frames}.
A popular dataset MultiWOZ~\citep{budzianowski-etal-2018-multiwoz} designed a user-friendly interface for the Wizards and easy-to-follow user goals, resulting in diverse and semantically rich data.
The recent Multi3WOZ~\citep{hu2023multi3woz} expands this paradigm, collecting a large-scale multilingual TOD dataset with parallel utterances in four languages over the same conversational flows.




\begin{figure*}[ht]
    \centering
    \includegraphics[width=\linewidth]{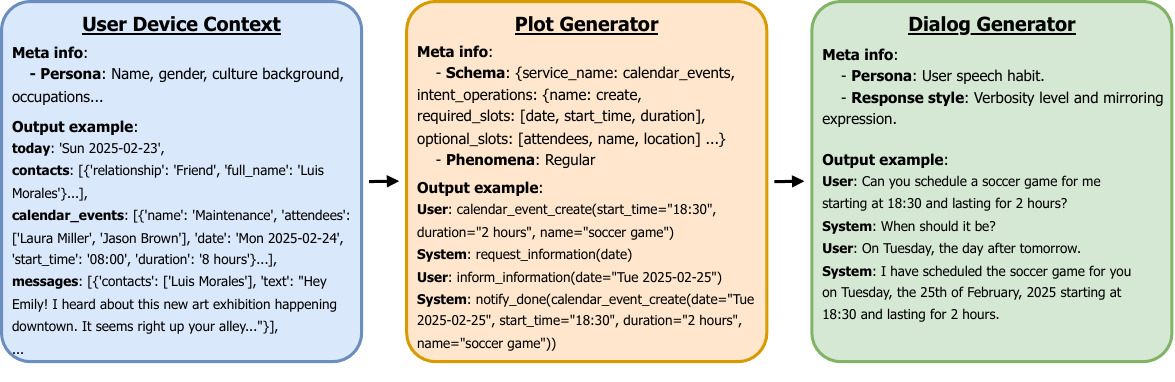}  

    \caption{Overview of the TOAD Automatic Generation Pipeline in 3 Steps: (i) Persona-grounded user device context generation, (ii) Action plot generation, and (iii) Dialog utterance realization.}
    \label{fig:overview}
    \vspace{-3mm}
\end{figure*}

\subsection{Dialog Distillation}
Recent advancements in LLMs have demonstrated their improved ability to follow human instructions and generate fluent conversation utterances for a range of understanding and generation tasks.
A list of recent works embrace this concept, extracting knowledge and dialog from LLMs to train student or in-domain models. 
SODA~\citep{kim2023soda} distilled social interactions dialogue by contextualizing commonsense knowledge from a social knowledge graph.
PLACES~\citep{Chen2023} generated a multi-party social dialog by prompting the OPT~\citep{zhang2022opt} model with topic, facts, and dialog as in-context learning examples.
MathDial~\citep{macina2023mathdial} distilled teacher-student math reasoning tutoring dialog through pairing human teachers with a LLM.
There are also works that distil multi-turn open-domain instructional conversations, such as Baize~\citep{xu2023baize} and UltraChat~\citep{ding2023enhancing}. 
Our TOAD represents the first work in distilling TOD dialog with multiple response style options, which requires more nuanced control over the conversation content.

\section{Data Generation Pipeline}
The TOAD pipeline, depicted in Fig.~\ref{fig:overview}, comprises three key stages for the automated creation of dialogues. 
The initial stage involves the generation of device context, simulating users' personas and their device app contexts. 
In the second stage, essential conversation attributes, including conversational content, flow, and language phenomena, are determined and incorporated into a dialog plot. 
The final stage involves the realization of the dialog plot into dialog utterances by a dialog generator. 
This generator simulates user speaking habits and provides a spectrum of system response styles.

\subsection{Persona-Grounded Context Generation}
Interacting with on-device context information such as calendar and alarms is an important part of real-life virtual assistant use-cases.
A recent dataset, PRESTO \citep{goel2023presto}, provides simulated structured contexts, but their influence is very limited as most dialogs are irrelevant to their contexts and there is no interaction such as modification or deletion.

Data diversity is important to the model's robustness and generalization.
However, LLM-generated data given the same input prompt often lacks diversity.
To address the two issues mentioned above, we develop a persona-grounded context generation pipeline, which combines sampling from external data sources and chain-of-thought generation~\citep{wei2022chain}. Recent work \citep{hu2024quantifying} has shown that LLMs are capable to simulate different perspectives.

We synthesize each persona in 3 steps:
1) We synthesize random occupation information, by sampling from work statuses including employed, unemployed/retired, and student. For students or the employed, occupations are based on NAICS 6-digit industries.\footnote{\url{https://www.census.gov/naics/2022NAICS/2022_NAICS_Structure.xlsx}} Additional occupation details such as location, affiliation, and job level are generated by prompting the LLM.
2) We sample surname and race together from the 2010 US Census data,\footnote{\url{https://www2.census.gov/topics/genealogy/2010surnames/names.zip}} and sample other attributes such as gender, MBTI personality from pre-defined sets. 
3) The sampled information is input into the LLM to write an introduction with additional fictionalized details (e.g. first name, age, marital status, hobbies).

Based on each persona's introduction, attributes and a random `current time', we prompt the LLM to generate app context instances for each service. Details about the supporting services are provided in Section~\ref{sec:stats}.
There are dependencies across services, e.g. information of generated contacts might be shared by messages and calendar.



\begin{table*}[]
\centering
\renewcommand{\arraystretch}{1}
\setlength{\tabcolsep}{5pt}
\scalebox{0.93}{
    \begin{tabular}{l p{8cm} p{6cm}}
        \hlinewd{1pt}
        \addlinespace[0.5ex]
        \textbf{Phenomena}        & \multicolumn{1}{l}{\textbf{Actions}}     & \multicolumn{1}{l}{\textbf{Utterance}}   \\ 
        \hlinewd{0.4pt}
        \addlinespace[0.5ex]
        \multirow{2}{*}{\small Compound}         & 
        \small \texttt{\textbf{restaurant\_booking\_reserve\_table}(restaurant ="French Brasserie", time="8:00 PM"),} & 
        \small I would like to reserve a table at the French Brasserie for 8:00 PM. Also, can you help me find a hotel in Las Vegas? \\ 
        \addlinespace[-2.5ex] & 
        \small \texttt{\textbf{hotel\_booking\_search\_hotel}(location="Las Vegas")}      &                        \\
        \addlinespace[0.5ex]
        \hlinewd{0.4pt}
        
        \small Compositional    & 
        \small \texttt{\textbf{weather\_get\_weather}(date=get\_calendar\_events( name="Art Class").\textbf{calendar\_events\_check}(date).date)}       & 
        \small What's the weather like on the day of my Art Class event?      \\ 
        \addlinespace[0.5ex]
        \hlinewd{0.4pt}
        
        \small Self-correction  & 
        \small \texttt{\textbf{get\_movie\_time}(movie\_name="Fast \& Furious Presents: Hobbs \& Shaw", location="Miami").self\_correction( location="Houston")}       & 
        \small Could you find the showtimes for Fast \& Furious Presents: Hobbs \& Shaw in Miami? Actually, make that Houston instead.       \\ 
        \addlinespace[0.5ex]
        \hlinewd{0.4pt}
        
        \small Complex referral & 
        \small \texttt{\textbf{get\_alarms}(ordered\_by="time", index=0).check(time)}       & 
        \small What's the time for my earliest alarm?      \\ 
        \addlinespace[0.5ex]
        \hlinewd{1pt}
    \end{tabular}
}
\caption{Dialog Phenomena Examples. Actionns and utterance for initial query examples, For multi-intent phenomena, compound and compositional, we concate service name as prefix to the intent actions.}
\label{tab:phenomena}
\vspace{-3mm}
\end{table*}

\subsection{Plot Generation}



\paragraph{Schema}
Our pipeline, inspired by the approach in SGD~\citep{rastogi2020towards}, establishes a schema for each service by outlining supported intents, associated slots and relevant meta-data. 
For example, the \textit{calendar\_events} service has four supporting intents (\textit{create}, \textit{modify}, \textit{check} and \textit{delete}).
Each intent specifies required slots like \textit{start\_time}, \textit{date} and \textit{duration} along with optional slots such as \textit{attendees}, \textit{name} and \textit{location}.
Additionally, we also specify a list of meta-indicators for each intent to control the conversation flow. 
For instance, the intent of \textit{reserve\_table} is marked as for \textit{``require\_confirmation=True''}, indicating the need for an additional confirmation in the dialog flow.
Further details about the schema are provided in Appendix~\ref{app:schema}.

\paragraph{Dialog Phenomena}
To enhance the authenticity of our conversation flows, we simulate various dialog phenomena, exemplified in Tab.~\ref{tab:phenomena}. 
These phenomena are categorized into multi-intent and single-intent types.
Compound and compositional dialogs fall into the multi-intent category, where users pursue related or unrelated goals simultaneously or their initial intent depends on another implicit intent.
In the Tab.~\ref{tab:phenomena} example, to obtain the weather information for the day of the ``Art Class'', an implicit intent \textit{check} is needed to get the actual date of the ``Art Class''.
Self-correction and complex referral are phenomena that simulate real-life conversation dynamics.
In self-correction, users correct themselves or alter their requests mid-sentence or afterward. 
In complex referral, users employ realistic referring expressions that require additional logic to comprehend.
For instance, an oral expression like ``earliest alarm'' requires ranking all alarms in app context by time to pinpoint the user's reference accurately.

\paragraph{Intent Sampling}
In TOAD, intent represents user's intention and determines the conversation content and goal.
To compose the dialog, we sample services and intents, and generate corresponding slot values through LLM prompts. 
For consistency, all slot values are generated simultaneously. 

For multi-intent compound dialog, two intents and their respective slot values are independently sampled. 
While for compositional dialog, as the output of the implicit intent will be used as the input of the initial intent, the two intents are jointly sampled with a slot matching strategy. 
Further details on the intents sampling and slot values generation are provided in Appendix~\ref{app:intent sampling}.

\paragraph{Plot Representation}
As illustrated in Fig.~\ref{fig:overview}, a plot serves as a conversation skeleton, consisting of a sequence of Meaning Representations (MRs) that encapsulate all the necessary dialog information. 
The manipulation of these MRs allows control over the dialog's content and flow. 
In contrast to SGD and STAR, which rely on a finite set of function-like actions for MRs, TOAD opts for a pseudocode format to represent actions. 
This choice avoids dependence on a fixed set of predefined actions, enriching the meaning space, particularly in terms of referring expressions. Furthermore, the decision to use pseudocode aligns with the fact that most latest LLMs have been pre-trained on code, ensuring their comprehension of code-like MRs.

\begin{figure}
    \centering
    \includegraphics[width=0.85\linewidth]{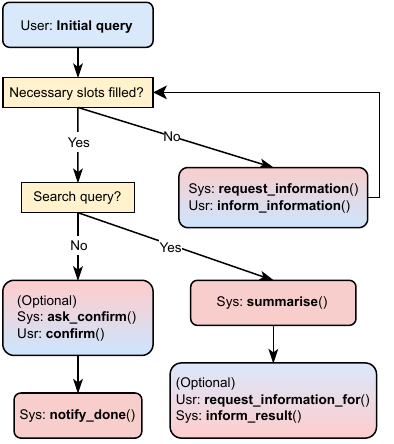}  
    \caption{Plot construction for single intent dialog based on slot-filling strategy. }
    \label{fig:plot}
    \vspace{-3mm}
\end{figure}

\paragraph{Plot Construction}
Dialog plot construction follows the flowchart in Fig.~\ref{fig:plot}. 
The core mechanism employs a slot-filling strategy, with the system asking the user to fill all required slots to achieve their intent. 
The meta-indicators specified in the intent schema also determine the plot components, e.g. including additional confirmation turns or replying with search result summarization.
For multi-intent dialogues, the plots are composed by merging individual plots for each intent in a natural order.
The details of the merging policy are provided in Appendix~\ref{appen:plot merging}.

\paragraph{Context Interaction}
Dialogues involving interactions with context apps are ubiquitous in real-life scenarios but have not been adequately addressed in existing datasets. 
TOAD is the first work that includes conversations with context interactions, supporting generic intents of checking, deleting, and modifying on app services such as calendar events, alarms, and messages. 

This expansion significantly broadens the scope for simulating dialogues with natural referring expressions and introduces the concept of proper referring expressions.
Users can refer to a calendar event using any slots, but the system should adhere to a preference order: 
While multiple slots may be unambiguous, certain slots are more intuitive for users. 
For instance, when referring to a calendar event, ``the 9 a.m. meeting'' is preferred over ``the meeting in the office.'' Therefore, referring by time is prioritized over referring by location. 
For each intent, we define a preference order for the slots, enhancing the naturalness of the system response.

\paragraph{Database query}
Traditional TOD datasets support intents that require database queries, such as restaurant or movie searching. 
In contrast to SGD's use of FreeBase for realistic slot values, our TOAD pipeline leverages LLMs as a knowledge database.
To ensure consistent query results, we prompt the LLMs with input slot values as conditions. 
For each query, we instruct LLMs to generate five different instances, simulating the search operation and ensuring diversity in the returned values.
The advantages of using LLMs as a knowledge base include: 1) It avoids the issues of empty or noisy searching results. 2) The query outcomes are more consistent with other slot values.



\subsection{Dialog Generator} \label{sec:dialog generator}
The dialog generator realizes the composed plots into natural utterances.
Similar to SGD and STAR, we adopt the Machine-to-Machine (M2M) approach, where different LLMs take turns to play the roles of user and system for generating conversation utterances. 

\paragraph{Style Control}
To offer natural responses across various scenarios, we generate a spectrum of response styles for each system turn as shown in the example in Fig.~\ref{fig:response style}, including combinations of low, mid and high verbosity and mirroring or no mirroring.
This is achieved by providing In-Context Learning~(ICL) examples and style definitions for different styles to LLMs.
The motivations behind these style designs are explained in Section~\ref{sec:response style}.

Given the six style options for each system turn, a selection must be made to construct the dialog history, which will shape the subsequent generation.
The dialog history significantly influences the generation of future conversation utterances. For instance, if the system consistently responds with low verbosity, user utterances tend to become shorter as well.
Hence, the most natural style for a given conversation context should be selected, referred to as the default style, which should balance between efficient communication and minimizing the risk of misoperation. 
The selection rules are explained in detail in Section~\ref{sec:response style}.

\paragraph{Prompt Template}
The prompt templates for user and system encapsulate three key pieces of information: 1) Current turn dialog action, 2) dialog history, and 3) style instruction. 
ICL demostration for action-to-utterance are also provided to guide the utterance realization.
For user turns, the style instruction describes the user's speaking habit based on the sampled persona. 
For system turns, the style instruction provides a brief definition of each style with corresponding ICL examples. 
Examples of prompt templates can be found in Appendix~\ref{app:dialog template}.

For efficiency and cost considerations, TOAD prompts LLMs to generate all system styles in a single inference pass.
While generating one style utterance at a time improves textual quality and control performance, it represents a trade-off between quality and cost.

\subsection{Dataset Quality Control}
To ensure the quality of the generated dataset, we conducted a two-step quality control process. The first step involved two automatic sanity checks: 1) Removing datapoints with utterances containing unformatted values, such as internal datetime representations like `2025-12-16'. 
2) Removing datapoints that have missing slot values in the corresponding utterances.

The second step addressed the potential inconsistencies between plot actions and utterances that may arise due to the non-deterministic nature of language models. 
Although human evaluation is considered the gold standard, it is not always feasible or scalable for large-scale dataset expansion. As an alternative, we propose using a language model evaluator to perform quality control, an approach that has gained popularity in recent works \citep{chen2023exploring,zeng2023evaluating,liu2024aligning}. These studies have demonstrated the effectiveness of language model evaluators in assessing various aspects of generation tasks, such as summarization and data-to-text. 

To implement this approach, we provide the LLM evaluator (gpt-3.5-turbo) with evaluation criteria, app context, action plot, and corresponding utterances. The evaluator then determines whether the utterances are consistent with the action plot and the app context. In our case, the LLM labeled 37 data points as inconsistent. Upon manual examination, we confirmed that 32 of these samples were indeed inconsistent. 
We listed the template used for quality control in Appendix~\ref{app:quality control template}.

\section{System Response Style}
\label{sec:response style}
The primary evaluation criterion for today's virtual assistants is their accuracy in responding and assisting users in achieving their goals. 
The next generation of virtual assistants is expected to exhibit increased naturalness and adaptability to various usage scenarios, which means the systems should be able to respond with different styles.
TOAD stands as the first work in investigating the dimensions of style that virtual assistants should consider.
We aim to open the discussion on optimal response styles to encourage future exploration.
In this work, our focus centers on two main aspects: verbosity level and mirroring. 

\subsection{Verbosity}
Verbosity refers to the degree of details or amount of information expressed in a response utterance. 
In many cases, clear and concise communication is preferred as it ensures efficient delivery and easy comprehension.
However, certain contexts may demand a higher verbosity to provide comprehensive explanations and avoid ambiguity. 
It's essential to strike a balance and use an appropriate level of verbosity based on the specific communication situation and user needs.

In TOAD, we define three levels of verbosity, illustrated through examples in Fig.~\ref{fig:response style}: Low verbosity responses should only contain a few words and may not necessarily be a complete sentence.
Mid verbosity responses should use pronouns to refer any previously mentioned events.
High verbosity responses should refer to events with proper and detailed expression and provide all relevant information.
System verbosity critically impacts user experience, and we propose a set of rules to determine the most natural response for a given context, which we consider the default style.

\begin{enumerate}[noitemsep,topsep=1pt]
    \itemsep 0em
    \item If there is special language phenomena in the user's last query, the system should employ higher verbosity.
    \item If the system needs to respond with new information, verbosity should be mid or high.
    \item If the system is requesting user confirmation for an irreversible operation, such as calender deleting or ticket booking, verbosity should be high.
    \item Otherwise, the system should exhibit low verbosity for efficient communication. 
\end{enumerate}
We note that this approach is subjective, and different styles may be favored in various situations.




\begin{figure*}[t]
    \centering
    \includegraphics[width=1\linewidth]{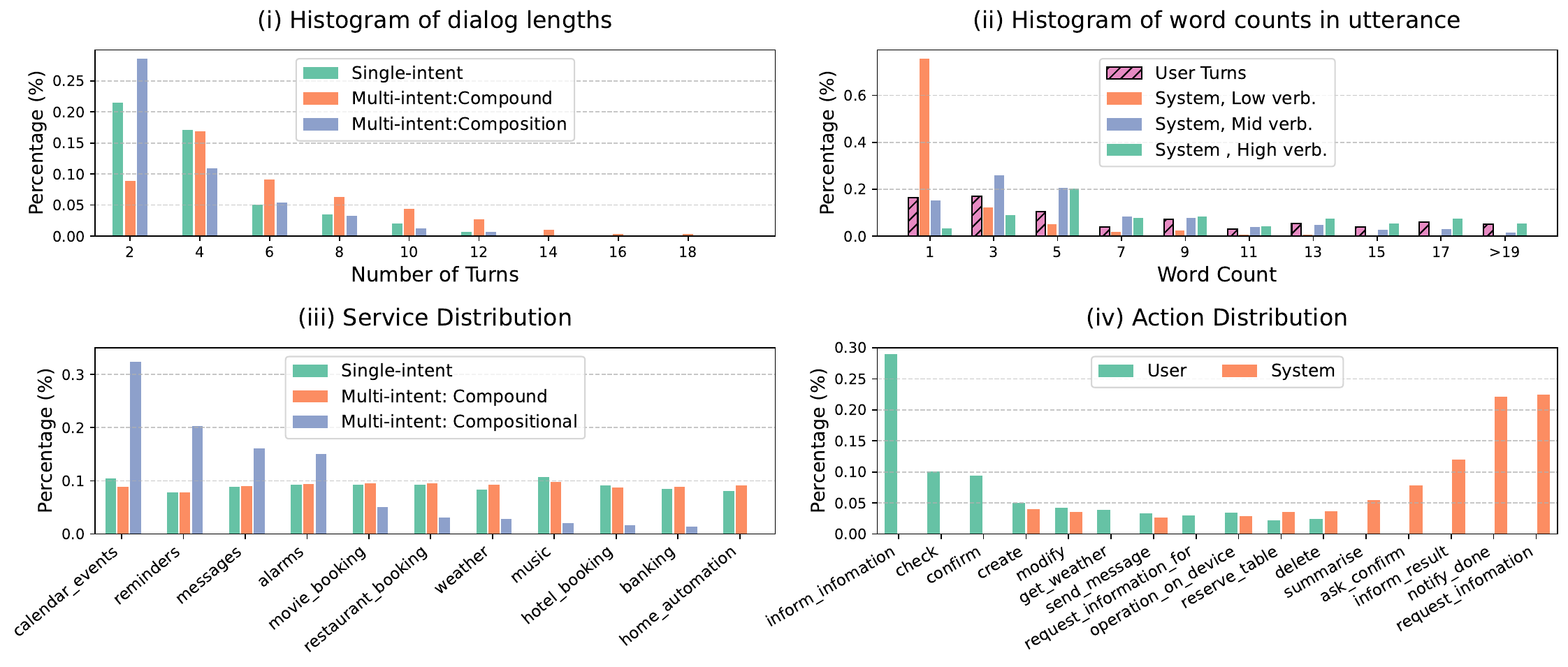}  
    \caption{(i) Distribution of the dialog lengths. (ii) Distribution of the word count in dialog utterance. (iii) Service Coverage Distribution in Dialogs. (iv) Distribution of plot actions (Actions with $>3\%$ inclusion). Note: All distributions are flattened, e.g. each service or action in multi-intent dialogs individually}
    \label{fig:distributions}
    \vspace{-3mm}
\end{figure*}

\subsection{Mirroring}
Extensive research \citep{liao2020racial,vartanov2023effect}
demonstrates the benefits of mirroring user behavior, emotions, personas, and expressions in psychological tasks. 
Mirroring enhances mutual understanding, promotes efficient communication, and consequently leads to improved user feedback~\citep{garrod2009joint}.
In TOAD, the response style of mirroring refers to the reuse of noun or verb phrases from the user's expression whenever feasible, as illustrated in Fig.~\ref{fig:response style}. 
In contrast, a non-mirroring style avoids directly copying the user's expression.

Mirroring is generally a favorable strategy, fostering natural conversation. 
However, caution is warranted, as a virtual assistant should consistently maintain professionalism and ensure positive communication.
Instances where mirroring may be inappropriate include:
\begin{itemize}[noitemsep,topsep=1pt]
    \itemsep 0em
    \item Emotional and improper phrasing: In real-life scenarios, users may use sentimental or inappropriate language, such as cursing words.
    \item Bias: Users' expressions may contain negative connotations related to gender, culture, race, or politics. A virtual assistant should avoid reinforcing these biases.
    \item Non-factual information: Users may not always provide accurate information, particularly in knowledge domains. For instance, asking ``Who is the king of the United States?''.
    \item Ambiguous terms or referring expressions: Mirroring potentially ambiguous expressions could be confusing to the user. 
    For example, a user might mention ``my meeting with David,'' and if the system copies this referring expression, the user could be uncertain when identifying the event, as they may not be aware of the possibility of having more than one scheduled meeting with David.

\end{itemize}
Mirroring user expressions is not universally safe or appropriate, and it is challenging to exhaustively enumerate all misuse cases.
The outlined situations provide a foundation for identifying potential challenges, and we encourage future exploration.

\section{Dataset Statistics}
\label{sec:stats}
The current TOAD dataset covers 11 service domains with 37,678 turns (18,839 user turn and 113,034 system response variations, 131,873 turns in total) in 8,087 dialogues.
It's worth highlighting that as TOAD operates as an automatic generation pipeline, both datasize and covered service can be scaled up.
Additionally, the creation of a new service schema typically requires only 15 minutes.

\begin{table*}[t]
\centering
\scalebox{0.95}{
    \begin{tabular}{lccccccccccccccc}
    \hlinewd{1pt}
    \multicolumn{1}{c}{\multirow{3}{*}{Model}} & \multicolumn{7}{c}{Act to Text} & \multicolumn{1}{l}{} & \multicolumn{7}{c}{DH+Act to Text} \\ \cline{2-8} \cline{10-16} 
    \multicolumn{1}{c}{}  & \multicolumn{3}{c}{Test} &  & \multicolumn{3}{c}{Zero-shot Test} & \multicolumn{1}{l}{} & \multicolumn{3}{c}{Test}  &  & \multicolumn{3}{c}{Zero-shot Test} \\
    \multicolumn{1}{c}{} & B  & R & M &       & B & R & M &       & B  & R  & M &       & B & R & M \\ \hlinewd{0.4pt}
    FlanT5-250m  & 44.8  & 61.8  & 65.0 &  & 28.4  & 51.5  & 55.1  &  & 54.2  & 67.2  & 69.4  &  & 38.3 & 55.3  & 59.3 \\
    FlanT5-3b  & \textbf{45.5} & \textbf{62.4} & \textbf{65.6}  &  & 34.2 & 54.3 & 56.9 & & 52.8 & 67.5 & 70.9  &  & 41.9  & \textbf{59.2} & \textbf{64.0} \\
    FlanT5-11b  & 43.0  & 60.9  & 64.2   &  & \textbf{35.9}  & \textbf{57.8} & \textbf{60.8}  & & \textbf{54.9} & \textbf{68.9} & \textbf{72.0}  &  & \textbf{44.7} & \textbf{59.2} & 63.3  \\
    Llama2-7b  & 41.4  & 61.1 & 63.5  &  & 31.3  & 52.2 & 54.0 &  & 48.2   & 62.9   & 65.1  &  & 40.6  & 54.6  & 61.5 \\
    Llama2-13b  & 41.4 & 59.6  & 64.8  &  & 34.8 & 55.5 & 57.5  &  & 49.4 & 64.0  & 68.7 &  & 42.7  & 56.0 & 62.0 \\ \hlinewd{1pt}
    \end{tabular}
}
\caption{Results for two benchmarks, Action to utterance and Dialog History~(DH)+Action to utterance, reported in BLEU(B), Rouge-L(R) and Meteor(M) scores. All models are fine-tuned on the train set and evaluated on test and zero-shot test sets.
}
\label{Tab:results}
\vspace{-2mm}
\end{table*}

Across the 8087 dialogs, 40.5\% are multi-intent compound, 20.9\% are compositional and 38.6\% are single intent.
The distributions of the number of turns and word counts in utterances are shown in Fig.~\ref{fig:distributions}(i) and (ii) respectively.
As for local phenomena, 19.0\% dialog are self-revision, 15.1\% have complex referral, 3.3\% have both and 62.6\% have no local phenomena.
Note that local phenomena can appear in both single and multi-intent dialog.

Fig.~\ref{fig:distributions}(iii) illustrates the service distribution which is relatively uniform for single intent and compound dialogs.
However, in compositional dialog, due to the joint sample requirement, certain services may have less matchable slots. 
Scaling up the supporting services in the schema can potentially expand the joint service sample pool.

As mentioned above, TOAD does not have a fixed set of plot actions.
Fig.~\ref{fig:distributions}(iv) displays the distribution of actions whose
occurrences are greater than 3\%.
Based on the selection rules for the default response style, 61.7\% of the system responses are default high verbosity, 28.8\% and 9.5\% are mid and low verbosity.
The average word counts for these three levels are 17.6, 9.0 and 2.5 respectively.




\section{Evaluation}
TOAD data supports a range of TOD tasks, including Intent Detection, Slot Labeling in Natural Language Understanding, Dialog or Action State Tracking~(D/AST) and Natural Language Generation~(NLG).
This paper establishes benchmarks for two response generation setups, aligning with TOAD's primary objective of producing more natural, realistic dialogues and diverse response styles.

\subsection{Response Generation Benchmarks}
The first benchmark setup is a traditionally surface realization setup, where a dialog action is aimed to convert into a natural language utterance~\citep{williams2007partially}
The second, following \citet{hu2023multi3woz}, is `oracle' language model setup, where both previous dialog history and current action are available to produce the target utterance.
For both setups, the target verbosity and mirroring option are given as part of the input.

We partition the TOAD dataset into three sets: a training set ($\approx$ 81.8\%), a test set (10\%), and a zero-shot test set ($\approx$ 8.2\%) containing dialogs exclusively related to the `banking' service.
The zero-shot test set aims to assess models' generalization ability to unseen services.

Each response style option is treated as an individual datapoint. 
The surface realization setup only includes non-mirroring response options, as it does not have access to previous dialog utterances.
Whereas, the oracle language model setup considers all six options. 
In total, the two setups have 56.5K and 113K datapoints respectively

We applied supervised fine-tuning~(SFT) to five baseline models from two LM families: Encoder-decoder models, including FlanT5-(base, XL and XXL)~\citep{chung2022scaling}, with parameter sizes of 250M, 3B and 11B respectively, and decoder-only models Llama2-(7b, 13b)~\citep{touvron2023llama}.
The SFT hyper-parameters are provided in Appendix~\ref{app:sft templates}.

We evaluate model predictions using three widely-used NLG metrics:
BLEU~\citep{papineni2002bleu}, ROUGE-L~\citep{lin2002manual} and METEOR~\citep{banerjee-lavie-2005-meteor} scores.
The results are reported in Tab.~\ref{Tab:results}. 
In both setups, we observed that performance is generally better on the test set compared to the zero-shot test set, and increasing model sizes can improve the performance on both sets.
However, the performance gaps between the test and zero-shot test sets become smaller for LMs with increasing sizes, which we believe is due to the increasing zero-shot generalization ability of larger LMs.

Additionally, we consistently observe that the oracle language model setup outperforms the surface realization setup, which means having extra information of the dialog history can steadily improve the response generation performance.


\begin{table}[t]
    \centering
\scalebox{0.9}{
    \begin{tabular}{lcclccc}
    \hlinewd{1pt}
    \multicolumn{1}{c}{\multirow{2}{*}{Model}} & \multicolumn{2}{c}{Mirror} &  & \multicolumn{3}{c}{Verbosity} \\ \cline{2-3} \cline{5-7} 
    \multicolumn{1}{c}{}                       & w.          & w./o.        &  & High      & Mid     & Low      \\ \hline
    FlanT5-250m                                  & 56.0        & 47.8         &  & 45.3      & 49.8    & 51.0    \\
    FlanT5-3b                                  &\textbf{59.4} & 49.0         &  & 45.8      & 49.9    & 52.6    \\
    FlanT5-11b                                 & 59.2   &\textbf{49.4}    &  & \textbf{46.4}      & \textbf{50.1}    & \textbf{53.2}     \\
    Llama2-7b                                   & 49.3        & 43.6         &  & 38.1      & 43.6    & 50.5     \\
    Llama2-13b                                  & 51.5        & 45.3         &  & 39.7      &   45.8   & 52.3     \\ \hlinewd{1pt}
    \end{tabular}
}
    \caption{Results for inferring mirroring and non-mirroring responses, reported in BLEU score. All LMs are fine-tuned with the train set under DH+Act to utterance setting. }
    \label{tab:mirroring}
    \vspace{-2mm}
\end{table}
\subsection{Predictions on Different Response Styles}
In this section, we explore the modeling difficulty posed by different response styles. 
In Tab.~\ref{tab:mirroring}, we compare the performances on mirroring and non-mirroring responses as well as the performances across different verbosity styles.
All results are evaluated on the predictions generated by the oracle language model setup on the test set. 

We consistently noting that the performance on mirroring responses is better than non-mirroring ones, as expected, given that mirroring allows the system to replicate user expressions directly from the dialog history, making it easier to predict.
We also observe that as verbosity increases, performance tends to decrease. This decline is likely attributed to lower verbosity resulting in shorter utterances with a reduced vocabulary, making them easier to model.

Another observation worth noticing is that, even for non-mirroring options, having access to dialog history can yield improved modeling performance, as evident when comparing the left-most column of Tab.\ref{Tab:results} with the second left-most column of Tab.\ref{tab:mirroring}.

\section{Conclusion}


In conclusion, our study explores the naturalness and adaptiveness of system responses for the next generation of TOD virtual assistants. We introduce TOAD, a dataset designed to train TOD systems for diverse verbosity levels, mirroring styles, and realistic app context interactions. 
Additionally, we present a cost-effective and scalable automatic data generation pipeline as a practical alternative to traditional human annotations. 
By addressing those critical gaps, we aim for TOAD to inspire future exploration in modeling and analyzing system response styles.

\section{Limitations}
\paragraph{Entities}
The names and entities are generated by ChatGPT, therefore some of them are figures from publications, such as movies or novels.


\paragraph{Difficult cases for style modeling}
When producing the six response styles, certain scenarios pose challenges to model desired styles, which means the difference between some styles might become less evident.
We listed some observations during TOAD's construction:

\begin{itemize}[noitemsep,topsep=1pt]
    \itemsep 0em
    \item In cases when the system needs to present new information to the user, such as reporting database search results, in order to provide necessary information, sometimes low verbosity option can be relatively lengthy.

    \item When the system is requesting value for a new slot, it usually struggles to mirror users, who may not have mentioned the required slots.
    
    \item The use of proper referring expressions in user interactions with contextual apps can result in responses that are very similar between mirroring and non-mirroring styles.
\end{itemize}

\section{Ethics and Risks}


To ensure no user is disadvantaged, a dataset for model training needs to represent a diverse range of users.
The distribution of persona attributes in the TOAD dataset is detailed in Appendix~\ref{app:persona stats}.

\bibliography{anthology,custom}
\bibliographystyle{acl_natbib}

\appendix

\section{Schema Details and Example}
The schema is defined in JSON format. For each service, there are a list of supporting intents and a list of slots. For each intent, there are lists of ``required\_slots'', ``optional\_slots'' and ``result\_slots''. 
There are also meta-indicators to control the plot construction for that intent.
\label{app:schema}
\begin{lstlisting}[style=json, caption={Schema example for intent "create" of service "calendar\_events" with parts of the relevent slots.}, label={lst:schema}]
{
  "service_name": "calendar_events",
  "intent_operations": [
    {
      "name": "create",
      "description": "Create a calendar event.",
      "require_input_values": true,
      "require_context": false,
      "require_confirmation": false,
      "return_list": false,
      "report_result": false,
      "check_on_input": false,
      "can_refer_to_input_slot": true,
      "minimum_input_slot_number": 1,
      "minimum_initial_slots": [],
      "summary_emphasis_slots": [],
      "required_slots": [
        "date",
        "start_time",
        "duration_time"
      ],
      "optional_slots": [
        "attendees",
        "name",
        "location"
      ],
      "result_slots": []
    }
  ],
  "slots": [
    {
      "name": "date",
      "description": "The date of the calendar event.",
      "potential_values": [],
      "alias": ["start_date"]
    },
    {
      "name": "start_time",
      "description": "The time of the calendar event.",
      "potential_values": [],
      "alias": ["time"]
    },
    {
      "name": "duration_time",
      "description": "The duration_time of the calendar event.",
      "potential_values": [],
      "alias": []
    }
  ]
}
\end{lstlisting}

\section{Data Structure}
Each datapoint contains a multi-turn dialog action plot and corresponding utterances, each system turn contains 6 styles of responses and an indicator of the default style. 
Each datapoint also contains a user persona, relevant app contexts and labels for services, intents and phenomena. 

\section{Intents and slot values sampling}
\label{app:intent sampling}

The intent sampling for single-intent dialog is straightforward. A service is randomly sampled first and then one of the supported intents is randomly sampled.
For compound dialog, two intents are sampled by this method.

As for compositional dialog, intents are sampled based on the slot matching strategy, because the output slot of the inner intent, $s_{output}^{inner}$, will be used as the input slot of the outer intent, $s_{input}^{outer}$. As shown in the compositional example in Tab.~\ref{tab:phenomena}, the output slot of the inner intent ``check'' is ``date'', which is passed to the outer intent ``get\_weather'' as the input slot.
The slot matching strategy does not require $s_{output}^{inner}$ and $s_{input}^{outer}$ to be exactly the same, instead, it only requires $s_{input}^{outer}$ being a super-set of $s_{output}^{inner}$.
For example, in the phrase ``Book an alarm at the time when the show begins,'' the $s_{input}^{outer}=$``time'' is the super-set of the $s_{output}^{inner}=$``showtime''. 
As shown in the schema example in Listing~\ref{lst:schema}, for each slot, its parent slot is assigned in ``alias''.

To ensure the consistency among the slot values, all values are generated by LLM within one inference. The input slot values are generated first, such that the output slot values can condition on them.

\section{Multi-intent plot merging}\label{appen:plot merging}
Compound dialog contains two intents. When constructing the plot for compound dialog, the plots for two individual intents are merged into one, but with certain merging rules to maintain the naturalness of the conversation flow. Here are the merging policy: 
\begin{itemize}[noitemsep,topsep=1pt]
    \itemsep 0em
    \item Initial user query: Concatenate the actions of two individual initial queries and realize them into one single user utterance.

    \item For system response: If both system responses require user information, arrange the responses into two subsequent turns. Otherwise, combine two system responses into a single turn, but re-ordered to keep confirmation responses first.

    \item For subsequent user turn: Only answer the system response. 
\end{itemize}

As for compositional dialog, the initial user queries will be merged by substituting the matching input slot of the outer intent with the inner intent action, as illustrated in Tab.\ref{tab:phenomena}. 
Then the rest of the plots will be rearranged that the system should always finish the plot for the inner intent first.

\section{Persona and App Context Statistics}
\label{app:persona stats}
In TOAD, we simulate 500 personas with diverse backgrounds, with one persona randomly assigned to each dialog as the user's information source. The gender distribution of the simulated personas is approximately 52\% male and 48\% female.

Examining the racial distribution, we find that 17\% are White, 10\% Hispanic or Latino, 6\% Black or African American, 4\% Asian or Pacific Islander, 4\% American Indian or Alaska Native, and 59\% are not specified. The job descriptions encompass a total of 67 distinct occupations categorized into four job levels: senior (18\%), intermediate (31\%), entry-level (22\%), and not-specified (29\%).

In terms of context app distribution, the dialogs are diversified with 17.9\% involving calendar events, 13.5\% messages, 13.5\% reminders, 13.1\% alarms, 8.9\% contacts, and 33.1\% having no specified context.

\section{SFT hyperparameters}\label{app:sft templates}
SFT batch sizes for FlanT5-250m, 3b, 11b are 16, 16, and 4, respectively. Llama2-7b and 13b have batch sizes set at 2. The learning rates are $3e-5$, and fine-tuning halts upon meeting the early stopping criterion, which is 5 consecutive increases in validation set loss.

\section{Plot and Dialog Examples} \label{appen:examples}
We show more dialog and action plot examples in Tab.~\ref{tab:examples}. 
\begin{table*}[h]
    \centering
    \begin{tabular}{|p{4cm}|p{5cm}|p{7cm}|} 
        \hline
        \multicolumn{1}{|c|}{App Context} & \multicolumn{1}{c|}{Action Plot} & \multicolumn{1}{c|}{Dialog Utterances} \\
        \hline

\begin{lstlisting}[style=json_tab]
{'today':'Wed 2025-04-16', 
'alarms':[{'time':'07:00', 'name':'Morning Workout', 'if_repeat':True}, {'time':'18:30', 'name':'Family Dinner', 'if_repeat':True}]}
\end{lstlisting}
 & 
\begin{lstlisting}[style=text_tt]
@User@: [get_alarms(ordered_by="time", index=0).alarms_delete(name), get_alarms(ordered_by="time", index=1).alarms_check(name)]
@System@: [alarms_ask_confirm(get_alarms(name="Morning Workout").delete(name)), alarms_inform_result(name="Family Dinner")]
@User@: alarms_confirm()
@System@: alarms_notify_done(get_alarms(name="Morning Workout").delete(name))
\end{lstlisting}
 &  
\begin{lstlisting}[style=text_tt]
@User@: Can you delete my earliest alarm and check the next one?
@System@: {
HV M: Are you sure you want to delete the 'Morning Workout' alarm? Your next alarm is 'Family Dinner' at 18:30.
HV no_M: Do you confirm the deletion of your earliest alarm, 'Morning Workout'? The following alarm is 'Family Dinner' at 18:30. 
MV M: You want to delete the 'Morning Workout' alarm, correct? Your next one is 'Family Dinner' at 18:30. 
MV no_M: Confirm to delete this alarm? The next one is at 18:30 for 'Family Dinner'.
LV M: Delete 'Morning Workout'? Next: 'Family Dinner' 18:30. 
LV no_M: Confirm deletion? Next: 18:30 'Family Dinner'. }
@User@: Yes, go ahead and delete it.
@System@: {
HV M: Your 'Morning Workout' alarm has been deleted. 
HV no_M: The earliest alarm has been deleted. 
MV M: 'Morning Workout' alarm is deleted. 
MV no_M: It's deleted.
LV M: Deleted. 
LV no_M: Done.}
\end{lstlisting}  
\\ \hline
\begin{lstlisting}[style=json_tab]
{'today':'Mon 2026-12-28'}
\end{lstlisting}
&  
\begin{lstlisting}[style=text_tt]
@User@: purchase_tickets(movie_name="Joker", date="Tue 2026-12-29", cinema_name="Regal Cinemas", ticket_quantity=1, movie_format="3d")
@System@: request_infomation(showtime)
@User@: inform_infomation(showtime="20:00")
@System@: ask_confirm(purchase_tickets(movie_name="Joker", ticket_quantity=1, date="Tue 2026-12-29", showtime="20:00", movie_format="3d", cinema_name="Regal Cinemas"))
@User@: confirm()
@System@: notify_done(purchase_tickets(movie_name="Joker", ticket_quantity=1, date="Tue 2026-12-29", showtime="20:00", movie_format="3d", cinema_name="Regal Cinemas"))
\end{lstlisting}
&
\begin{lstlisting}[style=text_tt]
@User@: I'd like to buy a 3D ticket for Joker at Regal Cinemas tomorrow, please.
@System@: {
HV M: What time would you like to watch Joker at Regal Cinemas tomorrow? 
HV no_M: What showtime do you prefer? 
MV M: When would you like to see it? 
MV no_M: When is the preferred time?
LV M: Showtime? 
LV no_M: Time? }
@User@: At 20:00.
@System@: {
HV M: So, you'd like to purchase a 3D ticket for Joker at Regal Cinemas on Tue 2026-12-29 at 20:00. Is that correct? 
HV no_M: Just to confirm, you want to buy a 3D ticket for the movie Joker at Regal Cinemas on Tue 2026-12-29 at 20:00, right?
MV M: You're buying a 3D ticket for Joker at Regal Cinemas tomorrow at 20:00. Correct? 
MV no_M: So, you're purchasing a ticket for Joker in 3D at Regal Cinemas tomorrow at 20:00. Is that right?
LV M: Confirm purchase? 
LV no_M: Confirm? }
@User@: Yes, that's correct.
@System@: {
HV M: Your 3D ticket for Joker at Regal Cinemas on Tue 2026-12-29 at 20:00 has been purchased.
HV no_M: The ticket has been successfully purchased.
MV M: I've bought your ticket for that movie at that time.
MV no_M: It's done, the ticket is bought.
LV M: Done. 
LV no_M: Purchased. }
\end{lstlisting}
\\ \hline
\end{tabular}
\caption{Datapoint examples showing app context, action plots and dialog utterances with all response style options.}
\label{tab:examples}
\end{table*}

\onecolumn
\section{Dialog Generator Template}\label{app:dialog template}
\begin{table}[H]
\centering
\begin{tabular}{p{\linewidth}}
\hline
Prompt template for user utterance generation
\\ \hline
\begin{lstlisting}[style=text_rm]
"""\
You are a smartphone user and you are testing your virtual assistant on your phone by \
engaging in a multi-turn conversations with it. 
Here is your personal introduction: {{ user_intro }}

Instructions:
1. You need to communicate with the assistant following the guidance of "actions".
2. Based on the introduction, think about what speech habit you should have and \
communicate with this pattern.
3. The "message" should have minimum words possible.
4. You must return in JSON format, following the provided examples.

The context information is: {{ context }}.

You are using these apps: {{ situation }}.

Begin conversation (you are identified as "user").
user: {"actions": ["hello()"], "message": "Hi."}
assistant: {"actions": ["offer_help()"], "message": "Hello, how can I help?"}
{% if dialog_history | length>0 %}{{ dialog_history }}{% endif %}

The "actions" you need to follow is {{ action[cur_turn] }}. What are you going to say next? \
{{ user_style_instruction }}
user:\
"""
\end{lstlisting}
\\ \hline
\end{tabular}
\caption{Prompt template example for the user role in the dialog generator}
\end{table}

\begin{table}[h!]
\centering
\begin{tabular}{p{\linewidth}}
\hline
Prompt template for system utterance generation
\\ \hline
\begin{lstlisting}[style=text_rm]
"""\
Instructions:
1. You are a virtual assistant. Your goal is to assist the user to accomplish their goal.
2. Your responses should strictly follow the given actions and be helpful, natural, \
professional and concise.
3. Your response should strictly follow the corresponding "actions".

The user is interacting with these apps: {{ situation }}.
                                    
Style instruction:
'verbosity_low': Your response must only have a couple of words, such as "when", \
"how long" or "done".
'verbosity_mid': Your response should be a concise but complete sentence and must \
replace the nouns or noun phrases mentioned by user with pronouns, such as "it", "that" and "its".
'verbosity_high': Your response should use full expressions with all the details.
'mirroring': Your response should use the user's noun phrase or verb expressions when possible.
'no_mirroring': Ignore all previous dialog. Do not affect by user expression.
'summary': Your response should be a brief report of the given summary.

Response Example:
assistant actions: {"verbosity_low mirroring": ["notify_done()"], \
"verbosity_low no_mirroring": ["notify_done()"], \
"verbosity_mid mirroring": ["notify_done(operation_on_device(\
operation="turn_off", device="heating", home_space="bedroom"))"], \
"verbosity_mid no_mirroring": ["notify_done(operation_on_device(\
operation="turn_off", device="heating", home_space="bedroom"))"], \
"verbosity_high mirroring": ["notify_done(operation_on_device(\
operation="turn_off", device="heating", home_space="bedroom"))"], \
"verbosity_high no_mirroring": ["notify_done(operation_on_device(\
operation="turn_off", device="heating", home_space="bedroom"))"]}
assistant: {"verbosity_low mirroring": "Turned off.", \
"verbosity_low no_mirroring": "Done.", \
"verbosity_mid mirroring": "I have turned off the heating in that room.", \
"verbosity_mid no_mirroring": "I have turned it off in that room.", \
"verbosity_high mirroring": "I have turned off the bedroom heating.", \
"verbosity_high no_mirroring": "Sure, I have turned off the heating in the bedroom."}

Conversation history:
{% if dialog_history | length>0 %}{{ dialog_history }}{% endif %}

New turn:
You should return in JSON format with 6 keys: ["verbosity_low mirroring", \
"verbosity_low no_mirroring", "verbosity_mid mirroring", "verbosity_mid no_mirroring", \
"verbosity_high mirroring", "verbosity_high no_mirroring"].
assistant actions: {{ action[cur_turn] }}.
assistant: \
"""
\end{lstlisting}
\\ \hline
    \end{tabular}
    \caption{Prompt template example for the system role in the dialog generator}
\end{table}


\newpage

\section{Quality Control Template}\label{app:quality control template}
\begin{table}[H]
\centering
\begin{tabular}{p{\linewidth}}
\hline
Prompt template for quality control
\\ \hline
\begin{lstlisting}[style=text_rm]
"""\
Please check the consistency between the actions and the corresponding utterances for the following dialog. They might refer to the context below.

Context: {{ context }}

Dialog: {{ actions_and_utterances  }}

Response: \
"""
\end{lstlisting}
\\ \hline
\end{tabular}
\caption{Prompt template example for examining the consistency between action plot and utterances}
\end{table}

\end{document}